\documentclass[12pt]{article}

\usepackage[utf8]{inputenc}
\usepackage[T1]{fontenc}
\usepackage{amsmath,amssymb}
\usepackage{graphicx}
\usepackage{booktabs}
\usepackage{array}
\usepackage{longtable}
\usepackage{hyperref}
\usepackage{url}
\usepackage{geometry}
\usepackage{authblk}
\usepackage{microtype}
\usepackage{parskip}
\usepackage{listings}
\usepackage{xcolor}
\usepackage{float}
\usepackage{caption}
\usepackage{subcaption}
\usepackage{natbib}

\lstdefinestyle{json}{
  basicstyle=\ttfamily\small,
  breaklines=true,
  frame=single,
  backgroundcolor=\color{gray!10},
}

\title{\textbf{TerraMARS: A Domain-Adapted Small-Language-Model Pipeline for\\
Mars Terraforming Literature}}

\author[1]{Jyotsna Singh\thanks{Corresponding author: \href{mailto:jyotsnasingh@arizona.edu}{jyotsnasingh@arizona.edu}}}
\author[1]{Ash Black}

\author[2]{Jeff Larsen}

\author[3,4]{Scott R. Saleska}

\affil[1]{College of Information Science, University of Arizona, Tucson, AZ, USA}
\affil[2]{Biosphere 2, University of Arizona, Tucson, AZ, USA}
\affil[3]{Department of Ecology and Evolutionary Biology, University of Arizona, Tucson, AZ, USA}
\affil[4]{Department of Environmental Sciences, University of Arizona, Tucson, AZ, USA}

\date{}

\definecolor{jsonkey}{rgb}{0.16,0.42,0.66}
\definecolor{jsonstr}{rgb}{0.40,0.55,0.13}
\definecolor{jsonnum}{rgb}{0.66,0.20,0.20}
\definecolor{jsonbg}{rgb}{0.97,0.97,0.97}

\lstdefinelanguage{json}{
    morestring=[b]",
    morestring=[d]',
    literate=
        *{0}{{{\color{jsonnum}0}}}{1}
         {1}{{{\color{jsonnum}1}}}{1}
         {2}{{{\color{jsonnum}2}}}{1}
         {3}{{{\color{jsonnum}3}}}{1}
         {4}{{{\color{jsonnum}4}}}{1}
         {5}{{{\color{jsonnum}5}}}{1}
         {6}{{{\color{jsonnum}6}}}{1}
         {7}{{{\color{jsonnum}7}}}{1}
         {8}{{{\color{jsonnum}8}}}{1}
         {9}{{{\color{jsonnum}9}}}{1}
}

\lstdefinestyle{json}{
    language=json,
    backgroundcolor=\color{jsonbg},
    basicstyle=\ttfamily\small,
    breaklines=true,
    showstringspaces=false,   
    frame=single,
    framerule=0pt,
    xleftmargin=1em,
    aboveskip=1em,
    belowskip=1em,
    stringstyle=\color{jsonstr},
    keywordstyle=\color{jsonkey}\bfseries,
    columns=fullflexible,
    keepspaces=true,
    morekeywords={true,false,null}
}
\begin{document}

\maketitle

\begin{abstract}
 Researchers are interested in learning about Mars so that it may eventually become habitable for humans. To achieve this, there is a need for comprehensive knowledge of the planet's atmosphere, hydrology, surface chemistry, radiation environment, and spatial features through the scientific literature. These contain valuable information and meaningful quantitative constraints that can be used in other models and studies, such as habitability assessment and future terraforming studies. We present TerraMARS, an end-to-end information extraction pipeline that combines a domain-adapted Small Language Model to answer Mars terraforming-related questions and convert unstructured Mars science text into machine-readable structured outputs in JavaScript Object Notation (JSON) format. A corpus of open-access papers is collected and processed using a multistage retrieval and chunking framework. Google Gemma 3 1B was adapted to the domain using Quantized Low-Rank Adaptation (QLoRA) fine-tuning on Mars-specific question-answering and information extraction datasets. The resulting pipeline generates both types of output and provides a foundation for integrating knowledge from scientific literature into downstream applications like digital twins and habitability modeling for Mars. The output from this pipeline looks promising, but further improvements are needed to increase extraction accuracy and factual consistency.
\end{abstract}


\section{Introduction}

\qquad Terraforming Mars has been a topic of research for many years, with various researchers proposing frameworks to make the lifeless planet habitable for plants and humans. McKay et al. (1991) argued that terraforming Mars might be easier for plants: an initial warming triggered by greenhouse gases such as CFCs would release CO$_2$ trapped in polar caps and regolith through a positive feedback mechanism, creating a thicker atmosphere capable of eventually supporting plant life. Creating an Earth-like atmosphere for humans, however, would be more difficult due to the requirement of nitrogen and the sustained need for artificial warming.  Zubrin and Mckay (1993) proposed technological approaches to greenhouse Mars, including using its own CO$_2$ feedback system via orbital mirrors, imported volatiles, or manufactured halocarbon gases, as well as redirecting NH$_3$-rich outer solar system asteroids using nuclear thermal rockets as a source of volatiles. These studies reported specific quantitative details, such as the dimensions and mass of the orbital mirror and other relevant values.

\qquad  Increasing interest in this field has produced multiple publications containing valuable quantitative information (DeBenedictis et al., 2025; Stork et al., 2025; Ansari et al., 2024), raising the need to extract such information in a structured format like JavaScript Object Notation (JSON). Researchers working in Mars terraforming must review the latest papers, yet the number published exceeds what they can read, and the task becomes harder still when a researcher is interested in particular terraforming constraints. This bottleneck delays Mars terraforming research, motivating automation, since manual extraction is time-consuming and prone to human errors such as duplication, incorrect values, and outdated information. Automated extraction of scientific information has precedent in other fields. Krallinger et al. (2017) reviewed methods for automatically extracting chemical information from text and converting chemistry knowledge into structured data, and Tshitoyan et al. (2019) proposed an unsupervised machine learning method to extract knowledge from 3.3 million materials science abstracts. Transformer-based language models followed, including Bidirectional Encoder Representations from Transformers (BERT) Devlin et al. (2019), a general-purpose transformer encoder trained on large-scale English text using masked language modeling and next-sentence prediction to learn bidirectional context; SciBERT, Beltagy et al. (2019), a variant of BERT trained on scientific papers; and PubMed BERT, Gu et al. (2021), a BERT variant trained only on biomedical literature such as PubMed.

\qquad  The release of Large Language Models (LLMs) has opened further avenues for improving research workflows. These models, however, have shown hallucination in structured output tasks such as JSON generation (Bechard and Ayala, 2024; Tonmoy et al., 2024). General LLMs are also general in nature, meaning they are not tailored to a specific field, and they are expensive, requiring substantial graphics processing unit (GPU) memory that is generally a challenge in academic settings. Parameter-efficient fine-tuning methods such as Low-Rank Adaptation (LoRA), Hu et al. (2022) and Quantized Low-Rank Adaptation (QLoRA), Dettmers et al. (2023) address this by enabling adaptation to specialized domains with modest computational resources. LoRA reduces trainable parameters by several orders of magnitude by adding small trainable low-rank matrices into the attention layers of a Transformer instead of updating the full model. QLoRA extends this by keeping the same LoRA adapters while further compressing the frozen base model to 4-bit precision using NormalFloat4 (NF4), making it possible to fine-tune models of up to 65B parameters on a single GPU. In this paper, we investigate whether a 1-billion-parameter open small language model can be adapted to generate answers and extract structured quantitative constraints in JSON format from Mars terraforming literature.

\section{Data Sources}

\qquad In the present study we have focused on the literature related to Mars surface conditions, atmospheric properties, Mars surface chemistry, regolith properties, water availability, microbial habitability, and proposed terraforming interventions.  We restrict the corpus to open-access publications whose temporal scope spans approximately three decades, from 1994 to 2026, to ensure a balance between historical foundational works and recent advances. Three complementary open-access repositories were selected: arXiv was chosen for its broad coverage of physics, planetary science, and computer science preprints; PubMed Central (PMC) for biomedical and life science literature relevant to astrobiology and microbial habitability; and Semantic Scholar for aggregated cross-publisher coverage that may surface papers not indexed in the other two sources. 

\section{Methodology}
\begin{figure}[H]
\centering
\includegraphics[width=1\linewidth]{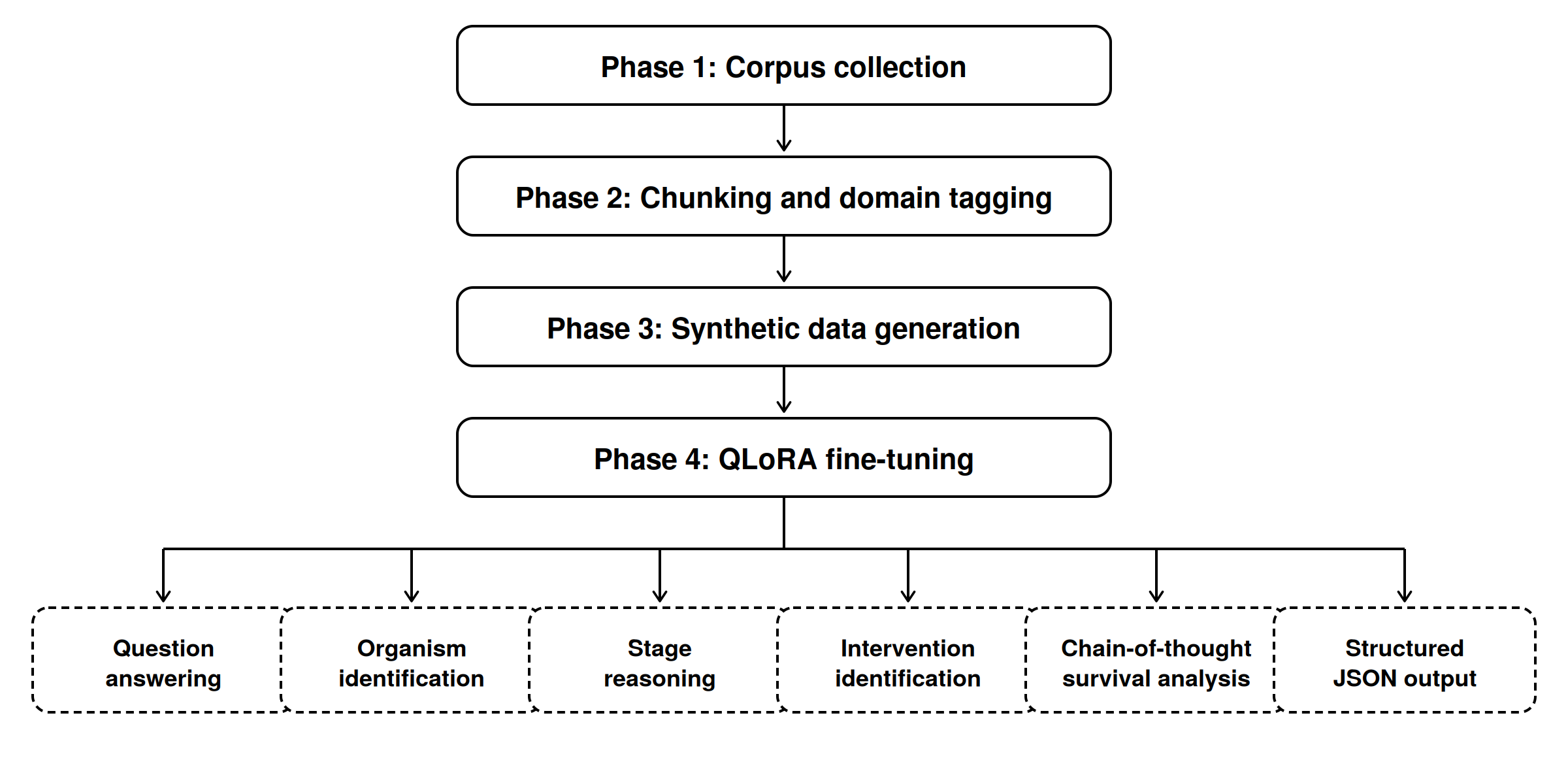}
\caption{TerraMARS methodology pipeline.}
\label{fig:pipeline}
\end{figure}

\qquad The methodology consists of four sequential phases: corpus collection, processing, synthetic data generation, and QLoRA fine-tuning. The resulting fine-tuned model is a single multi-task model that can produce six different output types selected by the prompt template at inference time: question answering, organism identification, stage reasoning, intervention identification, chain-of-thought survival analysis, and structured JSON extraction (Figure~\ref{fig:pipeline}). Each phase is described in the subsections below.

\subsection{Corpus Collection}

\qquad In the present study we have focused on the literature related to Mars surface conditions, atmospheric properties, Mars surface chemistry, regolith properties, water availability and microbial habitability. The retrieval process is guided by a set of 18 search queries designed to cover key scientific dimensions of Martian habitability and terraforming research (Table~\ref{tab:queries}). For each query, the metadata returned by the source Application Programming Interface (API)  is captured, and each paper is stored as an individual JSON record on disk. A minimum abstract length of 100 characters is enforced to exclude entries that contain only stub metadata. Papers whose abstract-level text cannot be retrieved by the source API are excluded at this stage. Duplicate records across the three sources are removed.
\begin{table}[H]
\centering
\caption{List of 18 search queries used for corpus retrieval}
\label{tab:queries}
\begin{tabular}{ll}
\toprule
\multicolumn{2}{c}{\textbf{Search query}} \\
\midrule
Mars terraforming                          & Extremophile Mars analog \\
Mars atmospheric pressure increase        & Mars perchlorate microorganism \\
Mars pioneer organisms' astrobiology      & Mars water activity microbial \\
Mars regolith biotransformation           & Mars soil formation organic carbon \\
Cyanobacteria Mars survival               & Deinococcus Mars radiation \\
Mars greenhouse gas warming               & Mars oxygen production photosynthesis \\
Mars nitrogen fixation                    & Mars permafrost subsurface life \\
Mars ISRU in-situ resource utilization    & Mars UV radiation microbial survival \\
Mars climate model habitability           & Mars sulfate brine habitability \\
\bottomrule
\end{tabular}
\end{table}

\subsection{Processing}

\qquad Each retrieved paper goes through a four-step processing pipeline. First, the raw abstract is cleaned by removing citation markers, hyperlinks, non-ASCII control characters, and extra whitespace; this normalization step prevents downstream tokenizers from treating formatting artifacts as semantic content. Second, the cleaned abstract is assigned to one or more of eleven domain categories using keywords (Table~\ref{tab:domains}). The eleven categories are: atmosphere, water, regolith, nitrogen, oxygen, pioneer biology, timeline, radiation, \textit{in-situ} resource utilization (ISRU), survival, and general. A chunk receives the label ``general'' only when no domain-specific keyword matches. The keyword dictionary is derived from eighteen search queries that combine the keyword ``Mars'' with domain-specific terms such as ``terraforming,'' ``atmospheric pressure,'' ``regolith,'' or ``UV radiation.'' Domain tagging is multi-label: a single chunk can be assigned to more than one category if its content matches keywords from multiple domains, which reflects the cross-disciplinary nature of Mars terraforming research. Third, the cleaned text is segmented into chunks of approximately 400 words. The 400-word chunk size is chosen to keep each chunk within the context window of the small language model used in downstream phases while preserving paragraph-level coherence. Because we operate on abstracts rather than full text, most papers produce a single chunk. Fourth, all chunks are written to a single JSONLines file in which each record contains the chunk identifier, the source paper's title and URL, the domain tags, the chunk index, and the chunk text. The JSONLines format is chosen because it supports streaming and incremental processing without loading the entire corpus into memory.

\begin{table}[H]
\centering
\caption{Details of eleven domains with keywords}
\label{tab:domains}
\begin{tabular}{>{\bfseries}p{3.5cm} p{9.5cm}}
\toprule
Domain & Keywords (case-insensitive) \\
\midrule
General         & Assigned when no category-specific keyword matches \\
Water           & water activity, brine, ice, subsurface water, permafrost, hygroscopic \\
Atmosphere      & atmospheric pressure, co2, greenhouse, climate model, global circulation, warming, temperature increase \\
Regolith        & regolith, soil formation, mineral weathering, basalt, perchlorate, iron oxide, jarosite, clay mineral \\
ISRU            & isru, in-situ resource, resource utilization, propellant production, sabatier, electrolyzer \\
Timeline        & terraforming timeline, centuries, millennia, stage, phase transition, habitability threshold \\
Nitrogen        & nitrogen fixation, nitrogenase, ammonia, n2, nitrogen cycle, nitrate \\
Oxygen          & oxygen production, photosynthesis, electrolysis, o2 partial pressure, oxygenic \\
Radiation       & uv radiation, uv flux, ionizing radiation, cosmic ray, radiation shielding, dna damage, ld50 \\
Pioneer biology & cyanobacteria, pioneer organism, lichen, biofilm, chlorella, anabaena, deinococcus, halophile, xerophile \\
Survival        & microbial survival, desiccation, freeze-thaw, sporulation, endospore, metabolic reactivation \\
\bottomrule
\end{tabular}
\end{table}

\subsection{Synthetic Data Generation}

\qquad In the third phase we generate synthetic training data using a ``teacher model'' because fine-tuning needs (prompt, answer) pairs and no pre-existing Mars instruction dataset exists. It is a common practice in machine learning known as knowledge distillation. We use Llama 3.2 3B served via Ollama as the teacher. The teacher reads each chunk and generates instruction-output pairs automatically across six task types: constraint extraction, terraforming stage reasoning, pioneer organism identification, intervention planning, quantitative question-and-answer pair generation, and chain-of-thought survival analysis.

\qquad For the structured extraction template, we additionally require that every generated constraint contain all five required fields (parameter, value, unit, condition, terraforming stage); any output that fails this strict schema check is discarded.

\subsection{Fine Tuning}

\qquad In the fourth phase we fine-tune Google Gemma 3 1B IT using QLoRA. The base model is loaded with 4-bit NF4 quantization and double quantization enabled. The dataset is then split into 95/5 training and validation set and converted into a HuggingFace dataset. Commonly adopted configurations were used---rank ($r$) equal to 16 and scaling factor ($\alpha$) equal to 32, applied to all seven attention and feed-forward projection matrices (\texttt{q\_proj}=query projection, \texttt{k\_proj}=key projection, \texttt{v\_proj}=value projection, \texttt{o\_proj}=output projection, \texttt{gate\_proj}=feed-forward gate projection, \texttt{up\_proj}=feed-forward up-projection, \texttt{down\_proj}=feed-forward down-projection). Training is conducted on Jetstream2 high-performance computing system (Hancock et al., 2021). We train for two epochs and used the AdamW optimizer (Loshchilov and Hutter, 2017) with a learning rate of $2 \times 10^{-4}$ and a cosine learning-rate schedule with a warmup ratio of 0.05. We use bfloat16 mixed-precision training and gradient checkpointing. The detailed hyperparameters used in the present model are given in Table~\ref{tab:hyperparams}.

\begin{table}[H]
\centering
\caption{Training hyperparameters}
\label{tab:hyperparams}
\begin{tabular}{ll}
\toprule
\textbf{Setting} & \textbf{Value} \\
\midrule
Base model              & google/gemma-3-1b-it \\
Quantization            & 4-bit NF4 with double quantization \\
LoRA rank ($r$)         & 16 \\
LoRA alpha ($\alpha$)   & 32 \\
LoRA dropout            & 0.05 \\
Target modules          & $^*$q\_proj, k\_proj, v\_proj, o\_proj, gate\_proj, up\_proj, down\_proj \\
Training examples       & 1,120 \\
Validation examples     & 59 \\
Epochs                  & 2 \\
Optimizer               & AdamW \\
Learning rate           & $2 \times 10^{-4}$ \\
LR schedule             & Cosine decay \\
Warmup ratio            & 0.05 \\
Effective batch size    & 32 \\
Gradient accumulation   & 16 \\
Max sequence length     & 256 \\
Precision               & bfloat16 \\
Training approach       & Supervised Fine-Tuning (SFT) with LoRA adapters \\
\bottomrule
\multicolumn{2}{p{12cm}}{$^*$q\_proj=query projection, k\_proj=key projection, v\_proj=value projection, o\_proj=output projection, gate\_proj=feed-forward gate projection, up\_proj=feed-forward up-projection, down\_proj=feed-forward down-projection,
LoRA- Low Rank Adaptation, LR- Learning Rate}
\end{tabular}
\end{table}

\section{Results and Discussion}

\subsection{ Corpus Distribution and Training Results}

\qquad The retrieval pipeline yielded 614 abstracts from arXiv (369), PMC (220), and Semantic Scholar (25). These sources were selected to cover the broad disciplines that can help us to understand Mars habitability. PMC entries were retrieved but excluded because abstract-level text was not returned by the NCBI esummary endpoint used in this pipeline. In addition, one arXiv entry was excluded because its abstract was below the 100-character minimum length and was filtered out; all retained abstracts therefore exceed 100 characters. After the quality control, we were left with 393 usable abstracts. These abstracts have been assigned 11 categories after cleaning; the chunk distribution is given in Table~\ref{tab:chunks}. Most of the chunks are in general category followed by water (88) and atmosphere (80) and least went into survival (3).

\qquad We constructed the datasets using synthetic instruction generation pipeline. Across the pipeline, a larger set of synthetic samples was generated that passed through filtering and validation. It led to 1179 high quality examples across six templates---constraint extraction (190), question-and-answer pair generation (199), pioneer organism identification (199), terraforming stage reasoning (210), intervention planning (186) and chain-of-thought survival analysis (195). The final pass rate was 99.7\%.

\begin{table}[H]
\centering
\caption{Distribution of chunks across the eleven domain categories used for chunk-level tagging}
\label{tab:chunks}
\begin{tabular}{lc}
\toprule
\textbf{Domain} & \textbf{Chunks} \\
\midrule
General         & 163 \\
Water           & 88 \\
Atmosphere      & 80 \\
Regolith        & 45 \\
ISRU            & 40 \\
Timeline        & 19 \\
Nitrogen        & 19 \\
Oxygen          & 18 \\
Radiation       & 15 \\
Pioneer biology & 9 \\
Survival        & 3 \\
\bottomrule
\end{tabular}
\end{table}

\subsection{TerraMARS Outputs}

\qquad We present six illustrative outputs from the fine-tuned model across the six prompt templates used during instruction tuning: question answering, organism identification, stage reasoning, intervention identification, chain-of-thought survival analysis and structured JSON extraction. The examples include both chunk-grounded responses where the template matches the chunk content and cases where the template forces a framing incompatible with the chunk. These outputs are illustrative and are not intended as quantitative evaluation.

\subsubsection{Question Answering}

\textit{Source chunk (excerpt) from Hu (2019), Predicted Diurnal Variation of the Deuterium to Hydrogen Ratio in Water at the Surface of Mars Caused by Mass Exchange with the Regolith:}

\begin{quote}
Regolith on Mars exchanges water with the atmosphere on a diurnal basis and this process causes significant variation in the abundance of water vapor at the surface\ldots{} We find that the D/H can vary by 300--1400 per mil diurnally in the equatorial and mid-latitude locations, and the magnitude is greater at a colder location or season. The variability is mainly driven by adsorption and desorption of regolith particles, and its diurnal trend features a drop in the early morning, a rise to the peak value during the daytime, and a second drop in the late afternoon and evening, tracing the water vapor flow into and out from the regolith.
\end{quote}

\textit{\textbf{4.2.1a Pipeline output}}

\begin{quote}
Q: How does the D/H ratio vary in the atmosphere of Mars?

A: The D/H ratio can vary by 300--1400 per mil diurnally in the equatorial and mid-latitude locations, and the magnitude is greater at a colder location or season. The variability is mainly driven by adsorption and desorption of regolith particles, and its diurnal trend features a drop in the early morning, a rise to the peak value during the daytime, and a second drop in the late afternoon and evening, tracing the water vapor flow into and out from the regolith.
\end{quote}

\textit{\textbf{4.2.1b Analysis}}

\qquad The model produces a chunk-grounded answer that shows the quantitative value (300--1400 per mil) present in the source chunk. The qualitative description of the diurnal pattern (morning drop, daytime peak, evening decline) is also taken directly from the chunk. This answer is fully supported by the source. Such variability is consistent with prior studies of Martian atmospheric fractionation processes, which show that isotopic enrichment in water vapor is influenced by surface--atmosphere exchange under diurnally varying temperature conditions (Villanueva et al., 2015).

\subsubsection{Organism Identification}

\textit{Source chunk (excerpt) from Turbet et al.\ (2020), Measurements and Semi-Empirical Calculations of CO$_2$+CH$_4$ and CO$_2$+H$_2$ Collision-Induced Absorption:}

\begin{quote}
Reducing atmospheres have recently emerged as a promising scenario to warm the surface of early Mars enough to drive the formation of valley networks\ldots{} We performed 1-D numerical radiative-convective climate calculations (using the LMD Generic Model) to compute the surface temperature expected on the surface of early Mars for several CO$_2$, CH$_4$, and H$_2$ atmospheric contents\ldots{} These calculations demonstrate that thick CO$_2$+H$_2$-dominated atmospheres remain a viable solution for warming the surface of Mars above the melting point of water, but not CO$_2$+CH$_4$-dominated atmospheres.
\end{quote}

\textit{\textbf{4.2.2a Pipeline output}}

\begin{quote}
Based on the provided text, the organism, microbial property, or biological adaptation being discussed is the CO2+CH4 collision-induced absorption (CIAs) of CO2 and CH4. The conditions affecting this property are the wavelengths and temperatures of the CIAs.
\end{quote}

\textit{\textbf{4.2.2b Analysis}}

\qquad This example illustrates a known failure mode of template-driven generation in small instruction-tuned models. The extracted chunk is about atmospheric chemistry and radiative transfer, not biology. However, the template prompt requires the model to identify ``an organism, microbial property, or biological adaptation'' in the text. The model attempts to satisfy the template by incorrectly framing CO$_2$+CH$_4$ collision-induced absorption as a ``microbial property.'' This behavior reflects template-domain mismatch: when the chunk is incompatible with the prompt template, the model generates a syntactically valid but semantically incorrect response.

\subsubsection{Stage Reasoning}

\textit{Source chunk (excerpt) from Kite and Daswani (2019), Geochemistry Constrains Global Hydrology on Early Mars:}

\begin{quote}
Ancient hydrology is recorded by sedimentary rocks on Mars\ldots{} We find that the CO$_2$ sequestration implied by the global groundwater hypothesis for the origin of sulfate-rich rocks on Mars is 30--5000 bars\ldots{} If, instead, Hesperian sulfate-rich rocks are approximated as pure Mg-sulfate (no Fe), then the CO$_2$ sequestration is 0.3--400 bars\ldots{} In both cases, carbon sequestration by global groundwater circulation actively works to terminate surface habitability, rather than being a passive marker of warm Earthlike conditions.
\end{quote}

\textit{\textbf{4.2.3a Pipeline output}}

\begin{quote}
This information applies to the early stage of terraforming, specifically the microbial pioneer phase. The text discusses the geochemical constraints on global hydrology on Early Mars, which is a key step in understanding the conditions under which life could have originated on the planet. The study of sulfate-rich rocks on Mars provides a crucial link between the geological record and the climate of early Mars. The study of sulfate-rich rocks on Mars is also important because it provides a way to test the hypothesis that the cations for Mars' Hesperian sulfates were derived from deep-sourced groundwater. The study of sulfate-rich rocks on Mars is also important because it provides a way to test the hypothesis that the cations for Mars' Hesperian sulfates were derived from upwelling of deep-sourced groundwater.
\end{quote}

\textit{\textbf{4.2.3b Analysis}}

\qquad This example shows two limitations. First, the given chunk does not mention terraforming stages or a ``microbial pioneer phase''; the model is using them to satisfy the template, which is a form of template-induced hallucination. Second, in the response one can see ``deep-sourced'' and upwelling of deep-sourced that shows near-verbatim repetition. This repetition represents neural text degenerations (occurring when maximization-based decoding strategies get stuck in repetitive loops). Holtzman et al.\ (2019) suggested nucleus sampling to solve this problem.

\subsubsection{Intervention Identification}

\textit{Source chunk (excerpt) from Rugheimer et al.\ (2026), The Goldilocks Problem for Detecting Water in Terrestrial Planets: Constraining Water Abundances in the Mid-IR with LIFE:}

\begin{quote}
We investigate how well the Large Interferometer for Exoplanets (LIFE) mission concept can detect habitable conditions on exoplanets through the presence of atmospheric water vapor as a proxy for surface oceans\ldots{} We simulate LIFE-like noise at spectral resolutions R = 50 and 100 using LIFEsim and perform Bayesian atmospheric retrievals to determine the technical requirements for LIFE to confirm habitability\ldots{} Detectable atmospheric water may imply surface oceans, as water is highly reactive and rapidly removed by surface mineral reactions.
\end{quote}

\textit{\textbf{4.2.4a Pipeline output}}

\begin{quote}
Based on the provided text, the engineering intervention mentioned is the use of the Large Interferometer for Exoplanets (LIFE) mission concept to detect habitable conditions on exoplanets through the presence of atmospheric water vapor as a proxy for surface oceans.
\end{quote}

\textit{\textbf{4.2.4b Analysis}}

\qquad The model correctly identifies the LIFE mission concept as the intervention discussed in the chunk and accurately paraphrases its purpose (detecting habitable conditions on exoplanets through atmospheric water vapor). The response is short; it shows the model neither forced to answer anything about Mars terraforming nor invented any other details. All the provided answers are in the source and produced accurate summaries.

\subsubsection{Chain-of-Thought Survival Analysis}

\textit{Source chunk (excerpt) from Galletta et al.\ (2010), Bacterial Survival in Martian Conditions:}

\begin{quote}
\ldots{} we designed and built two simulators of Martian environments where to perform experiments with different bacterial strains: LISA and mini-LISA. Our LISA environmental chambers can reproduce the conditions of many Martian locations near the surface\ldots{} Inside LISA we have studied the survival of several bacterial strains and endospores. We verified that the UV light is the major responsible of cell death. Neither the low temperature, nor the pressure, nor the desiccation or the atmospheric changes were effective in this sense. We found that some Bacillus strains have a particular capability to survive for some hours in Martian conditions without being screened by dust or other shields. \ldots{} Samples covered by these dust grains have shown a high percentage of survival, indicating that under the surface dust, if life were to be present on Mars in the past, some bacteria colonies or cells could still be present.
\end{quote}

\textit{\textbf{4.2.5a Pipeline output}}

\begin{quote}
Step 1: The two hypotheses about the origin of life on Earth and Mars are that life originated from a unique progenitor on Earth or from a Lithopanspermia (Mars to Earth or vice versa).

Step 2: LISA and mini-LISA are designed to simulate Martian environments where to perform experiments with different bacterial strains.

Step 3: We have studied the survival of several bacterial strains and endospores in Martian conditions.

Step 4: We verified that the UV light is the major responsible of cell death.

Step 5: Neither the low temperature, nor the pressure, nor the desiccation or the atmospheric changes were effective in this sense.

Conclusion: Under the Martian surface, some bacterial strains have a particular capability to survive for some hours in Martian conditions without being screened by dust or other shields.
\end{quote}

\textit{\textbf{4.2.5b Analysis}}

\qquad This example is the strongest chunk-grounded output among the free-form examples. Each reasoning step references content explicitly present in the source chunk, and no external numerical values are introduced. The model correctly identifies UV light as the primary cause of cell death, the inertness of other Martian environmental factors, and the role of the LISA simulators. However, the conclusion phrase ``Under the Martian surface'' introduces a conceptual shift from the source, which describes survival under ``dust-shielded'' surface conditions rather than true subsurface habitation. This represents a mild overgeneralization where experimental shielding effects are reframed as subsurface environmental conditions.

\subsubsection{Structured Extraction (JSON) Output}

The structured extraction task is constrained by a strict JSON schema (parameter, value, unit, condition and terraforming stage), which can lead to occasional invalid or incomplete outputs from the language model. Here we are showing the outputs that passed the full validation.

\textit{Source fragment (excerpt) from Ramachandran et al.\ (2021), Experimental Investigation of the Atmosphere-Regolith Water Cycle on Present-Day Mars:}

\begin{quote}
\ldots{} when the temperature is above 273.15 K (0\,$^{\circ}$C), this frost can transform into liquid water, which can persist for up to 3.5 to 4.5 h at Martian surface conditions \ldots{} A mixture of these salts in a 10\% concentration with simulant produces an aggregated granular structure with a water gain of approximately 18- to 50-wt\%.
\end{quote}

\textit{\textbf{ 4.2.6a Pipeline output}}

\begin{lstlisting}[style=json]

{
  "constraints": [
    {
      "parameter": "temperature",
      "value": 273.15,
      "unit": "K",
      "condition": "above",
      "terraforming_stage": 0
    },
    {
      "parameter": "pressure",
      "value": 1,
      "unit": "atm",
      "condition": "above",
      "terraforming_stage": 0
    },
    {
      "parameter": "water_content",
      "value": 10,
      "unit": "%, not specified",
      "condition": "above",
      "terraforming_stage": 0
    },
    {
      "parameter": "water_gain",
      "value": 18,
      "unit": "%, not specified",
      "condition": "above",
      "terraforming_stage": 0
    },
    {
      "parameter": "water_gain_time",
      "value": 3.5,
      "unit": "h",
      "condition": "above",
      "terraforming_stage": 0
    },
    {
      "parameter": "water_gain_time_max",
      "value": 4.5,
      "unit": "h",
      "condition": "above",
      "terraforming_stage": 0
    }
  ]
}
\end{lstlisting}
\textit{\textbf{4.2.6b Analysis}}

The model produces schema-conformant JSON output with all five required fields. In this example, six constraints are extracted. Line-by-line inspection against the source chunk revealed some interesting facts about TerraMARS.

\begin{itemize}
  \item \textit{temperature} = 273.15 K, condition `above': fully grounded. Direct match to ``when the temperature is above 273.15 K (0\,$^{\circ}$C)'' in the source.
  \item \textit{pressure} = 1 atm: not present in the source chunk. The text references ``Martian surface conditions'' but does not state any value for atmospheric pressure. This value is an external assumption inserted by the model.
  \item \textit{water\_content} = 10\%: weakly supported. The chunk says ``a mixture of these salts in a 10\% concentration with simulant,'' which describes the salt fraction of an experimental mixture, not a water content constraint. The 10\% appears in the source but the parameter label and the treatment as a habitability constraint are not supported.
  \item \textit{water\_gain} = 18 (wt\%): partially grounded. The source states ``approximately 18- to 50-wt\%.'' The model has stored only the lower bound, discarding the upper bound of the range.
  \item \textit{water\_gain\_time} = 3.5 h: partially grounded. The source states ``persist for up to 3.5 to 4.5 h.'' The model has stored only the lower bound.
  \item \textit{water\_gain\_time\_max} = 4.5 h: fully grounded. The model correctly captures the upper bound of the same range.
\end{itemize}

Out of the six constraints, two are fully grounded, two are partially grounded , one is weakly supported and one is not present in the chunk at all. Overall, we can say the JSON shows strong format compliance with weaker content fidelity.

\section{Limitations}

\qquad It is important to discuss the limitations of the present work as this will help to make the future TerraMARS versions better. In the beginning we planned to build a minimum viable product, so instead of working with the full-length papers we have focused only on abstracts, as a result, our corpus was small. Another important observation that the abstracts are concise and usually leave out some important information and quantitative details. Using full-text papers would yield more chunks and increase the accuracy of retrieved content, which would eventually reduce hallucination.

\qquad We also observed a skewed domain distribution: the majority of chunks were tagged `general' (163), meaning they were nonspecific, followed by `water' (88). This led to underrepresentation of several critical Mars terraforming topics, which affected the quality of answers. The keyword dictionary used for automatic domain tagging was general in nature and needs to be properly selected after deeper research. Another point to mention is the use of the Gemma 3 1B model, which is small, has limited reasoning capacity, and a higher chance of hallucination. The training data contains only 1,179 synthetic examples generated by another small model (Llama 3.2 3B, used as the teacher), so there is a high chance that teacher errors transfer to the student.

\qquad Structured JSON extraction from the pipeline has a low success rate. Even when JSON output is schema-valid, individual field values are not always grounded in the source chunk. Finally, in some cases the model exhibits sentence-level repetition, a known degeneration mode of small autoregressive models under greedy decoding.

\section{Conclusion and Future Work}

\qquad In the present work, we present TerraMARS, a domain-specific tool designed to support researchers working on Mars terraforming research. This system enables users to find answers curated from Mars terraforming-focused literature and returns quantitative constraints in machine-readable JSON format. It enables researchers to quickly obtain answers to their queries and find thresholds for terraforming research without manually searching papers. The current system performance is limited by the size of the underlying corpus. The performance of TerraMARS could potentially be improved by scaling the corpus and making the corpus more diverse. Future scraping should address the issue of empty abstracts by selecting domain specific keywords.  Integration of retrieval-augmented generation may further support evidence-based answers. Future model performance could also be improved by choosing a higher-capacity Gemma model.

\section{Release}

\qquad The scripts can be found in the given repository: \url{https://github.com/jyotsnasingh11217/TerraMARS.git}.

\section{Acknowledgements}

\qquad The authors thank the University of Arizona for supporting this work. Computational resources were provided by the Jetstream2 high-performance computing system, which enabled model fine-tuning.

\section{References}

\qquad Ansari, Samaneh, et al.\ ``Feasibility of keeping Mars warm with nanoparticles.'' \textit{Science Advances} 10.32 (2024): eadn4650.

\qquad B\'echard, Patrice, and Orlando Marquez Ayala. ``Reducing hallucination in structured outputs via Retrieval-Augmented Generation.'' \textit{arXiv preprint arXiv:2404.08189} (2024).

\qquad Beltagy, Iz, Kyle Lo, and Arman Cohan. ``SciBERT: A pretrained language model for scientific text.'' \textit{Proceedings of the 2019 Conference on Empirical Methods in Natural Language Processing and the 9th International Joint Conference on Natural Language Processing (EMNLP-IJCNLP)}. 2019.

\qquad DeBenedictis, Erika Alden, et al.\ ``The case for Mars terraforming research.'' \textit{Nature Astronomy} 9.5 (2025): 634--639.

\qquad Dettmers, Tim, et al.\ ``Qlora: Efficient finetuning of quantized llms.'' \textit{Advances in Neural Information Processing Systems} 36 (2023): 10088--10115.

\qquad Devlin, Jacob, et al.\ ``Bert: Pre-training of deep bidirectional transformers for language understanding.'' \textit{Proceedings of the 2019 Conference of the North American Chapter of the Association for Computational Linguistics: Human Language Technologies, Volume 1 (Long and Short Papers)}. 2019.

\qquad Galletta, Giuseppe, Giulio Bertoloni, and Maurizio D'Alessandro. ``Bacterial survival in Martian conditions.'' \textit{arXiv preprint arXiv:1002.4077} (2010).

\qquad Gu, Yu, et al.\ ``Domain-specific language model pretraining for biomedical natural language processing.'' \textit{ACM Transactions on Computing for Healthcare (HEALTH)} 3.1 (2021): 1--23.

\qquad Hancock, David Y., et al.\ ``Jetstream2: Accelerating cloud computing via Jetstream.'' \textit{Practice and Experience in Advanced Research Computing 2021: Evolution Across All Dimensions}. 2021. 1--8.

\qquad Holtzman, Ari, et al.\ ``The curious case of neural text degeneration.'' \textit{arXiv preprint arXiv:1904.09751} (2019).

\qquad Hu, Edward J., et al.\ ``Lora: Low-rank adaptation of large language models.'' \textit{ICLR} 1.2 (2022): 3.

\qquad Hu, Renyu. ``Predicted diurnal variation of the deuterium to hydrogen ratio in water at the surface of Mars caused by mass exchange with the regolith.'' \textit{Earth and Planetary Science Letters} 519 (2019): 192--201.

\qquad Kite, Edwin S., and Mohit Melwani Daswani. ``Geochemistry constrains global hydrology on Early Mars.'' \textit{Earth and Planetary Science Letters} 524 (2019): 115718.

\qquad Krallinger, Martin, et al.\ ``Information retrieval and text mining technologies for chemistry.'' \textit{Chemical Reviews} 117.12 (2017): 7673--7761.

\qquad Loshchilov, Ilya, and Frank Hutter. ``Decoupled weight decay regularization.'' \textit{arXiv preprint arXiv:1711.05101} (2017).

\qquad McKay, C.P., Toon, O.B., and Kasting, J.F. (1991). Making Mars habitable. \textit{Nature}, 352(6335), 489--496.

\qquad Rugheimer, Sarah, et al.\ ``The Goldilocks Problem for Detecting Water in Terrestrial Planets: Constraining Water Abundances in the Mid-infrared with LIFE.'' \textit{The Astrophysical Journal} 1003.1 (2026): 85.

\qquad Stork, Devon, and Erika DeBenedictis. ``An Introduction to Mars Terraforming, 2025 Workshop Summary.'' \textit{arXiv preprint arXiv:2510.07344} (2025).

\qquad Tonmoy, S.M., et al.\ ``A comprehensive survey of hallucination mitigation techniques in large language models.'' \textit{arXiv preprint arXiv:2401.01313} (2024).

\qquad Tshitoyan, Vahe, et al.\ ``Unsupervised word embeddings capture latent knowledge from materials science literature.'' \textit{Nature} 571.7763 (2019): 95--98.

\qquad Turbet, Martin, Christian Boulet, and Tijs Karman. ``Measurements and semi-empirical calculations of CO$_2$+CH$_4$ and CO$_2$+H$_2$ collision-induced absorption across a wide range of wavelengths and temperatures. Application for the prediction of early Mars surface temperature.'' \textit{Icarus} 346 (2020): 113762.

\qquad Vakkada Ramachandran, Abhilash, Mar\'ia-Paz Zorzano, and Javier Mart\'in-Torres. ``Experimental investigation of the atmosphere-regolith water cycle on present-day mars.'' \textit{Sensors} 21.21 (2021): 7421.

\qquad Villanueva, G.L., et al.\ ``Strong water isotopic anomalies in the martian atmosphere: Probing current and ancient reservoirs.'' \textit{Science} 348.6231 (2015): 218--221.

\qquad Zubrin, Robert, and Christopher McKay. ``Technological requirements for terraforming Mars.'' \textit{29th Joint Propulsion Conference and Exhibit}. 1993.

\end{document}